*Башмур К.А., Тынченко В.С., Бухтояров В.В., Сарамуд М.В.*

# РОБОТ-ПРИНТЕР ДЛЯ СОЗДАНИЯ ЭЛЕМЕНТОВ ТЕХНОЛОГИЧЕСКОГО ОБОРУДОВАНИЯ ДЛЯ ПРОИЗВОДСТВА КОМПОНЕНТОВ БИОТОПЛИВНЫХ КОМПОЗИЦИЙ

Сибирский федеральный университет, Красноярск

Данное исследование посвящено поиску новых научно-технических решений в области возобновляемых источников энергии, в частности биотоплива. Биомасса является основным видом топлива для «зеленой» энергетики, на нее приходится две трети выработанной из возобновляемых источников энергии. Дальнейшее развитие отрасли зависит от совершенствования применяемых в ней техники и технологий. На примере очистного аппарата показана и опробована новая технология прототипирования его деталей с помощью роботизированного модуля. Использование пластиков в качестве деталей технологического оборудования является современным трендом и может обуславливаться низкой прочностью сцепления различных веществ с поверхностью этих деталей ввиду плохой смачиваемости и низких значений поверхностной энергии данных материалов по сравнению с металлами.

**Ключевые слова:** робот-принтер, 3D-печать, интеллектуальное управление, создание трёхмерных объектов

*Bashmur K.A., Tynchenko V.S., Bukhtoyarov V.V., Saramud M.V.*

# ROBOT-PRINTER FOR CREATING ELEMENTS OF TECHNOLOGICAL EQUIPMENT FOR THE PRODUCTION OF COMPONENTS OF BIOFUEL COMPOSITIONS

This study is devoted to the search for new scientific and technical solutions in the field of renewable energy sources, in particular biofuels. Biomass is the main fuel for green energy, accounting for two thirds of the energy produced from renewable sources. The further development of the industry depends on the improvement of the equipment and technologies used in it. On the example of a cleaning apparatus, a new technology for prototyping its parts using a robotic module is shown and tested. The use of plastics as parts of technological equipment is a modern trend and may be due to the low adhesion strength of various substances to the surface of these parts due to poor wettability and low values of the surface energy of these materials compared to metals.

Немногим более десяти лет назад трудно было представить, что с помощью трёхмерной печати можно будет создавать не только пространственные геометрические фигуры, но и реальные детали и технологические объекты. Технологиисоздания 3D-принтеров стремительно развиваются и сегодня уже можно свободно приобрести относительно недорогой3D-принтер, позволяющий воплощать в жизнь практически любой замысел конструктора.На смену металлообрабатывающим станкам приходят 3D-принтеры.

Современные 3D-принтеры формируют реальные трёхмерные предметы при помощи нити из специального ABSпластика. Существуют также модели принтеров, формирующие предметы из гипса, древесной нити, биоразлагаемогоPLAпластика [1]. Кроме того, разработаны системы, создающие предметы при помощи лазера путём спекания порошкового полимера. Изделия из ABSпластика получили широкое распространение благодаря высокой прочности, доступности и сравнительной лёгкости обращения с материалом. Детали, изготовленные из ABSпластика успешно заменяют, например, зубчатые колёса во всевозможных кинематических парах, лопасти пропеллеров и другие.

Одним из наиболее распространённых для широкого пользования, в настоящее время, является 3D-принтер RepRap (рис. 1). В отличие от других принтеров, конструкция и программное обеспечение принтераRepRap[2], разработанного в Университете Бата (Великобритания), доступны на условиях открытого исходного кода. Одними из главных преимуществ данного принтера являются простота конструкции и возможность распечатки деталей для самостоятельной сборки ещё одного принтера-клона.

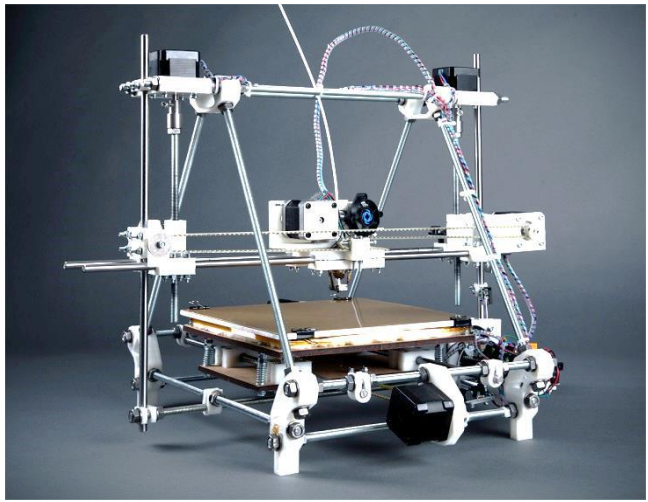

Рис. 1 – 3D-принтер RepRap

Принтер RepRap может работать с как с ABS, так и с PLA пластиком. Максимальные размеры изделия, которое можно распечатать на данном принтере - 200х200х200 мм. При этом достигаемая достаточно высокая точность воспроизведения размеров: 0.1-0.2 мм [2].

Ближайший конкурент RepRap - это 3D-принтер конструкции Delta (рис. 2), также достаточно простой по конструкции и эффективный [3]. Механизм позиционирования печатающей головки (экструдера) данного принтера базируется на механике дельта-робота. Он состоит из трехрычагов, прикрепленных посредством карданных шарниров к основанию. Это позволяет добиваться высокой скорости печати при малом количестве составных частей и быстро заменять головку экструдера. Принтер характеризуется более высокой, чем RepRapобластью печати: максимальный размер изделия 200х200х280 мм.Во время его работы движется только печатающая головка, а сам объект остается неподвижным.Рабочая платформапринтера снабжена стеклянной поверхностью и нагревательным элементом, что позволяет использовать в качестве материала для печати как PLA, так и ABS пластик.

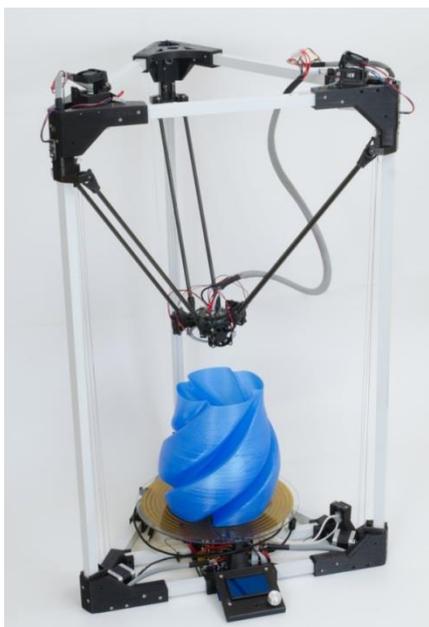

Рис. 2- 3D-принтер Delta

Основные современные методы создания технологических объектов при помощи трёхмерной печати, которые также можно применять в описанных выше базовых конструкциях, представлены в таблице 1.

Необходимо отметить, что основным недостатком описанных конструкций являются ограничения по габаритным размерам печатаемого изделия, регламентируемые размерами рамной конструкции самого принтера.Увеличение размеров печатной камеры ведёт к пропорциональному уменьшению полезного пространства и снижению точности позиционирования печатающей головки.

Применение 3D-принтеров в современном исполнении требует особого технологического подхода к организации его рабочего места, настройки оборудования и технологии печати[4]. Необходимо тщательно соблюдать температурные режимы, кинематическую точность направляющих,которая пропорционально уменьшается с размерами изделия, так как возникают паразитные прогибы. Всё это ограничивает развитие машины в области искусственного интеллекта. Другими словами, стационарный принтер в современных условиях, требующих более высокой точности при работе сосложными технологическими объектами, не обеспечивает должный уровень гибкости и свободы перемещения исполнительных механизмов и устройств.

Кроме того, повышенные требования к точности обуславливают возникающие требования ксамообслуживанию, самонастройке и калибровке машины. Человеческий фактор при этом сводится к минимуму: роль человекасводится только лишь к тому, чтобы сформулировать задание и нажать на кнопку. Поэтому необходимо модернизировать не только конструкцию принтера, но и систему управления.

Авторами предлагается вместо стандартной рамы и шаговых двигателейможно использоватьробот-манипулятор (рис.3), в который установить печатающую головку, как например, разработка французских дизайнеров Galatea, предназначенная для трёхмерной печати мебели [5]. Таким образом, экструдер будет иметь пять степеней свободы, а приводы современных роботов достаточно точны, для обеспечения точного позиционирования печатающей головки в пространстве.

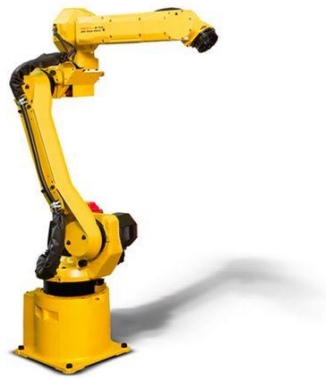

Рис. 3- Робот-манипулятор

Однако применение робота-манипулятора полностью не решит проблему увеличения пространственной свободы принтера. Его также необходимо установить на подвижное шасси, например, на колёсном ходу (рис.4). Это обеспечит свободное перемещение экструдера в пространстве, благодаря чему исчезнут какие-либо ограничения на размер формируемого изделия.Ограниченная область действия манипулятора компенсируется возможностью перемещения принтера на необходимое расстояние. Кроме того, появляется возможность создавать («распечатывать») по ходу движения опорную конструкцию и перемещаться по ней на необходимую высоту для продолжения работы принтера.

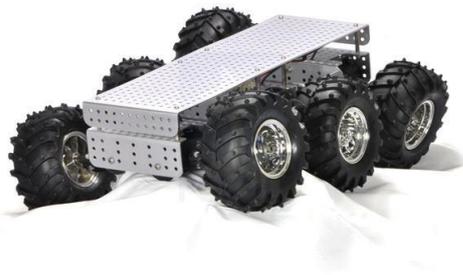

Рис. 4- Колесное шасси

Робот- принтер должен с высокой точностью определять своё местоположение, самостоятельно оценивать окружающее пространство. Такой принтер требует сложной системы управления и позиционирования, которую необходимо строить на принципах обратной кинематики и нечёткой логики. Такой подход в управлении будет первым шагом для перехода к нейронным сетям, а в последующем и к искусственному интеллекту.

Таблица 1 – Современные методы создания деталей

| Наименование | Описание |
|---|---|
| FDM (FusedDepositionModeling) | Моделирование методом осаждения расплавленной нити. |
| Polyjet | Основан на выпускании струи жидкого фотополимера, который образует слои на модельном лотке. |
| SL (Stereo Lithography) | Создание детали путем полимеризации жидкого полимера в рабочей ванне УФ излучением. |
| LS (LaserSintering) | Создание детали путем спекания порошка лазером. |
| LENS (Laser Engineered Net Shaping) | Использование энергии оптоволоконного лазера и металлической пудры, которая слой за слоем напыляет металлические объекты. |
| LOM (laminatedObjectManufacturing) | Объект формируется послойным склеиванием тонких плёнок рабочего материала, с вырезанием соответствующих контуров на каждом слое. |
| 3DP (ThreeDimensionalPrinting) | Послойное построение физических объектов за счет порошков со связующим материалом, наносимых последовательными тонкими слоями. |
| CJP (ColorJetPrinting) | Аналог печати 3DP с возможностью печати цветных объектов. |

В целом сборка такого робота-принтера не займёт много времени и позволит создавать большие объёмные модели технологических объектов и детали.

Процесс получения биодизельного топлива происходит в ходе химической реакции переэтерификации растительных масел. Полученное биодизельное топливо содержит примеси, такие как глицерин и твердые частицы, которые необходимо отделить, так как степень очистки биотоплива влияет на его энергетические характеристики и надежность технических систем. Существуют электростатические, центрифужные и другие технологии для отделения примесей. При этом используемые в данных технологиях аппараты достаточно громоздки, сложны в эксплуатации, имеют низкую пропускную способность и эффективность сепарации. Этих недостатков лишены гидроциклонные аппараты.

В лаборатории Биотопливных композиций Сибирского федерального университета разработаны устройства гидроциклонов, которые призваны решить вышеуказанные проблемы.

Для одной конструкции гидроциклона были изготовлены две втулки с рельефной винтовой поверхностью из полилактида с помощью робота-принтера (рисунок 5). Для другой конструкции гидроциклона был распечатан дисковый отражатель, представленный на рисунке 6.

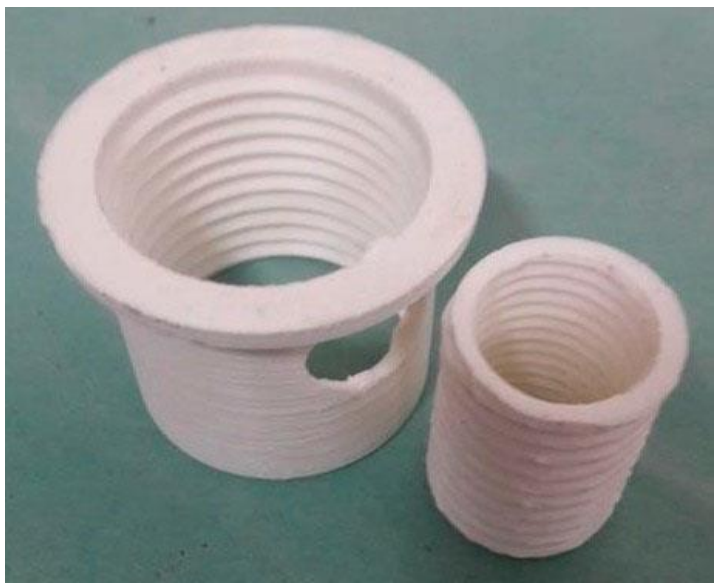

Рис. 5 – Распечатанные втулки для гидроциклона-демпфера

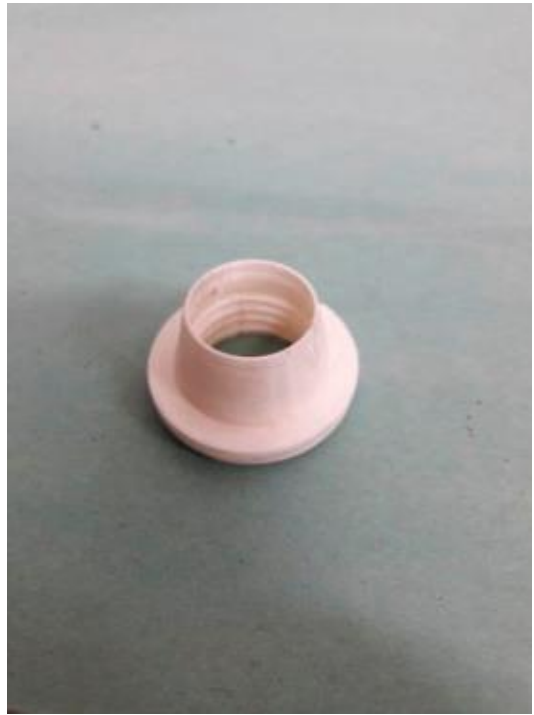

Рис. 6 – Распечатанный дисковый отражатель гидроциклона

Полилактид (PLA-пластик) является одним из наиболее экологически чистых материалов, применяемых для трехмерной печати, и таким образом практически идеально подходит для проведения экспериментов ввиду свойства биоразлажения.

Следует отметить, что использование пластиков в качестве деталей оборудования является современным трендом и, в частности, может обуславливаться низкой прочностью сцепления углеводородов с поверхностью этих деталей ввиду плохой смачиваемости и низких значений поверхностной энергии данных материалов, то есть снижением интенсивности и вероятности образования отложений на поверхностях деталей.

Управление роботом-принтером может осуществляться различным образом, но, пожалуй, наибольшего эффекта можно буде ожидать в случае использования системы управления на основе интеллектуальных технологий анализа данных. К ним в частности относят системы на нечеткой логике, искусственные нейронные сети и нейронечеткие системы. В задачах управления динамическими процессами нейронная сеть используется для выполнения нескольких функций. Нейронная сеть может выступать в качестве нелинейной модели процесса и идентифицировать его основные параметры, используемые при выработке соответствующего управляющего сигнала. Сеть может также играть роль следящей системы, отслеживающей условия окружающей среды и адаптирующейся к ним. Она также может выполнять функции нейрорегулятора, заменяющего собой традиционные устройства управления.

**Дополнительные сведения**

## СПИСОК ЛИТЕРАТУРЫ

1. Токарев Б.Е., Токарев Р.Б. Анализ рынка 3D-печати: технологии и игроки // Практический маркетинг. – № 2(204), 2014. – С. 10-16.
2. J.M. Pearce. RepRap for science – How to use, design and troubleshoot the self-replicating 3-D printer // Open-Source Lab. – Chapter 5, 2014. – P. 95-162.
3. Пушкарев В.В., Дроботов А.В. Компоновка устройств для объёмной печати экструдируемым расплавом деталей сложной формы // Известия Волгоградского государственного технического университета. - № 20 (123), 2013. – С. 121-123.
4. Данилов Р. Практический обзор 3D-принтера RostockBIV1 [Электронный ресурс]. –URL: http://3dwiki.ru/prakticheskij-obzor-3d-printera-rostock-bi-v1/
5. Сергиенко О. СтартапDrawn – распечатайте вашу мебель на роботизированном 3D-принтере![Электронный ресурс]. – URL:

http://3dwiki.ru/startap-drawn-raspechatajte-vashu-mebel-na-robotizirovannom-3d-printere/